\documentclass[letterpaper]{article} 
\usepackage{aaai23}  
\usepackage{times}  
\usepackage{helvet}  
\usepackage{courier}  
\usepackage[hyphens]{url}  
\usepackage{graphicx} 
\usepackage{natbib}  
\usepackage{caption} 
\usepackage{algorithm}
\usepackage{algorithmic}

\usepackage{booktabs}
\usepackage{amsmath}
\usepackage{pifont}
\usepackage{soul}
\usepackage{multirow}
\usepackage{newfloat}
\usepackage{listings}
\DeclareCaptionStyle{ruled}{labelfont=normalfont,labelsep=colon,strut=off} 
\floatstyle{ruled}
\newfloat{listing}{tb}{lst}{}
\floatname{listing}{Listing}
\title{Active Learning for Finely-Categorized Image-Text Retrieval\\ by Selecting Hard Negative Unpaired Samples}
\author{
Dae Ung Jo\quad
Kyuewang Lee\quad
JaeHo Chung\quad
Jin Young Choi
}

\affiliations{
    {\textrm{\{mardaewoon, kyuewang, jaehochung, jychoi\}@snu.ac.kr}}\\
    \vspace{1mm}
    Department of ECE, ASRI, Seoul National University, Korea
}

\begin{document}

\maketitle

\begin{abstract}

Securing a sufficient amount of paired data is important to train an image-text retrieval (ITR) model, but collecting paired data is very expensive. To address this issue, in this paper, we propose an active learning algorithm for ITR that can collect paired data cost-efficiently. Previous studies assume that image-text pairs are given and their category labels are asked to the annotator. However, in the recent ITR studies, the importance of category label is decreased since a retrieval model can be trained with only image-text pairs. For this reason, we set up an active learning scenario where unpaired images (or texts) are given and the annotator provides corresponding texts (or images) to make paired data. The key idea of the proposed AL algorithm is to select unpaired images (or texts) that can be hard negative samples for existing texts (or images). To this end, we introduce a novel scoring function to choose hard negative samples. We validate the effectiveness of the proposed method on Flickr30K and MS-COCO datasets.

\end{abstract}

\section{Introduction}
\label{section:introduction}
To increase the applicability of deep learning networks in various machine learning tasks, it is essential to collect sufficient amount of high-quality data for target applications.
However, collecting sufficient amount of data is a cost-consuming task.
Active learning (AL) is one of the methods to collect labeled data cost-efficiently.
AL assumes a situation where the annotator can annotate only a small number of data among a large amount of unlabeled data.
Among the given unlabeled data, AL algorithm selects limited number of samples to be annotated for cost-efficient training of the target model.
The actively selected samples by AL can further improve the performance of the target model, compared to the randomly selected samples.

In multi-modal applications, collecting data is more cost-consuming than single-modal applications.
In terms of reducing annotator's labor, AL for multi-modal tasks can be much more beneficial than AL for single-modal tasks.
However, many previous AL studies have mainly focused on single-modal tasks such as image classification~\cite{choi2021vab_vabal, 2019_learning_loss, 2019_active_vaal, choi2021vab_vabal,2018_active_coreset, ash2019BADGE} and AL for the multi-modal applications have not been explored much yet.
In this paper, we focus on AL for multi-modal applications, especially on image-text retrieval (ITR), which is one of the most popular multi-modal tasks.

In ITR, given a query image, an ITR model should retrieve relevant text from a database, and vice versa for a query text.
To train a model for ITR, most methods usually employ a contrastive learning scheme that leads a model to yield high similarity for relevant image-text pairs and low similarity for irrelevant pairs.
Thus, the training data for ITR contains lots of \textit{relevant image-text pairs} and their \textit{category labels} to predict relevance for other image-text pairs in the dataset.

In previous AL methods for ITR, annotators provide category labels for queried pairs~\cite{2018_alitr_coslaq}.
However, according to recent ITR studies~\cite{2017_itr_vsepp,2018_itr_scan,2020_itr_imram,2019_itr_camp,2019_itr_pvse,2019_itr_vsrn,2018_itr_scan}, category label becomes less important in the training phase.
The recent ITR studies have targeted challenging benchmarks, where relevant pairs are finely-categorized into numerous categories and each category contains very few pairs~\cite{2014_mscoco, 2014_flickr}.
Thus, during a training phase, most relevant pairs in a training mini-batch have different categories from others.
For this reason, many ITR studies assume that each relevant pair in the training dataset has its own category different from the categories of the other pairs, which does not need to compare its category label to the others.
Therefore, the category label is no longer utilized in the training phase and so asking a category label to annotators is meaningless for finely-categorized benchmarks.

In this paper, first, we set up an AL scenario that is feasible to finely-categorized ITR benchmarks.
In our scenario,
relevant pairs without category labels are used instead of category-annotated data.
Thus, an unpaired image is regarded as an unlabeled sample that is used for a query sample to request its paired text from the annotator.

For our AL scenario, we develop an AL algorithm of which key idea is to select unpaired images that are expected to produce a large training loss at a training phase. Samples causing a high loss can be regarded as hard samples to the current model. Thus, the model trained with the hard samples can achieve better performance than the model trained with randomly selected samples.
To this end, we utilize the triplet ranking loss function adopted in recent ITR studies that emphasize the hard negative samples~\cite{2017_itr_vsepp}.
Then we design an AL algorithm that selects images that can be a hard negative for as many texts as possible from the paired dataset.
To determine a hard negative image for a text in the paired dataset, we suggest a scoring function to measure the `hard negativeness' of each unpaired image sample for the given texts.
Our AL algorithm selects image samples in the order of the highest score.
Through extensive experiments on the Flickr30K~\cite{2014_flickr} and MS-COCO~\cite{2014_mscoco}, we validate the effectiveness of the proposed AL algorithm.

The contribution of the paper is summarized as follows.
\begin{itemize}
    \item[-]{We set up an AL scenario that is feasible to finely-categorized ITR benchmarks. In the scenario, a set of unpaired images is given and annotators provide paired texts for the images selected by an AL algorithm.}
    \item[-]{We propose an AL algorithm for our AL scenario, which can cost-effectively construct paired data beneficial for training the model to perform ITR tasks.}
    \item[-]{We validate the proposed AL algorithm through extensive evaluation and self-ablation studies on the Flickr30K and MS-COCO.}
\end{itemize}

\section{Related Works}
\label{section:related_works}

\subsection{Image-Text Retrieval Datasets}
\label{subsection:related_retrieval_dataset}
ITR dataset contains relevant image-text pairs categorized into several classes.
In datasets such as Wikipedia~\cite{rasiwasia2010new}, LabelMe~\cite{oliva2001modeling}, Pascal VOC2007~\cite{everingham2010pascal} and NUS-WIDE~\cite{chua2009nus} utilized in the previous literature~\cite{2018_alitr_coslaq}, data are categorized according to high-level (coarse) semantics. Therefore, each category contains many relevant pairs. We referred to those datasets as coarsely-categorized datasets.
But recent ITR studies have validated their algorithm at more challenging datasets such as Flickr~\cite{2014_flickr} and MS-COCO~\cite{2014_mscoco}. These datasets, on the other hand, contain data that are categorized according to low-level (fine) semantics. Thus, data samples are discriminated more thoroughly and precisely than the coarsely-categorized datasets. In other words, the number of category has increased, but the number of data samples has relatively decreased compared to coarsely-categorized datasets.
Table~\ref{tabular:data_config_citr_fitr} shows the configuration of the popular coarsely-categorized datasets~\cite{2018_alitr_coslaq} (top side) and finely-categorized datasets (bottom side).

\subsection{Image-Text Retrieval Methods}
\label{subsection:related_retrieval_method}
VSE++~\cite{2017_itr_vsepp} is the most popular ITR algorithm for finely-categorized benchmarks. VSE++ extracts features from image and text, and then calculates the cosine similarity between features. Then, a retrieval model is trained by the hinge-based triplet ranking loss~\cite{karpathy2015deep}.
However, when training a model, a mini-batch from Flickr or MS-COCO includes one relevant sample and a number of irrelevant samples. Thus, gradient can be biased to the negative term of the triplet loss. To mitigate this problem, VSE++ calculates loss function with only the hard negative sample in the mini-batch.
SCAN~\cite{2018_itr_scan} improves the retrieval performance by estimating the similarity between image and text more precisely. Compared to that VSE++ extracts the one global feature from image and text data, SCAN extracts lots of local features from sub-regions of an image~\cite{2018_bottomup} and words in a text.
Then SCAN calculates the similarity between each region (or word) and full text (or image) and finally aggregates them. Such algorithms are referred to as fine-grained ITR algorithms.
IMRAM~\cite{2020_itr_imram} refines local features by iteratively fusing local features with the proposed memory units.
In addition to the aforementioned studies, numerous ITR studies have been proposed~\cite{2019_itr_camp,2019_itr_pvse,2019_itr_vsrn,2018_itr_scan, zhang2022show, zhang2022negative, li2022action, diao2021similarity}.

\begin{table}[t]
\centering
        \begin{tabular}[h]{l|ccc}
            \toprule
            Dataset & \#Image & \#Text & \#Class\\
            \midrule
            NUS-WIDE-1.5K
            & 1,521  & 1,521  & 30\\
            LabelMe
            & 2,688  & 2,688  & 8\\
            Wikipedia
            & 2,866  & 2,866  & 20\\
            Pascal-VOC
            & 6,146  & 6,146  & 20\\
            \midrule
            Flickr8K
            & 8,091 & 40,455    & 8,091\\
            Flickr30K
            & 31,014 & 155,070    & 31,014\\
            MS-COCO
            & 123,287 & 616,435 & 123,287\\
            \bottomrule
        \end{tabular}
\caption{Configuration of the popular ITR benchmarks. 
}
\label{tabular:data_config_citr_fitr}
\end{table}

\subsection{Multi-Modal Active Learning}
\label{subsection:related_activelearning}
Recent AL studies have applied AL to various deep learning applications such as image classification~\cite{2018_active_coreset,2019_active_vaal,choi2021vab_vabal}, object detection~\cite{2019_active_detection}.
However, few studies have applied AL to multi-modal tasks.
COSLAQ~\cite{2018_alitr_coslaq} is an AL algorithm for coarsely categorized ITR datasets based on \citeauthor{2015_kang_citr}'s method~\cite{2015_kang_citr}. Given two relevant image-text pairs, COSLAQ calculates intra-modal (image to image, text to text) and cross-modal (image-text) similarities. The algorithm measures the variance of similarities and selects two relevant image-text pairs that have the highest similarity variance between them. Then COSLAQ queries human annotators whether two pairs belong to the same class or not.
\citeauthor{rudovic2019multi_mmql} proposed an AL algorithm for multi-modal data classification based on reinforcement learning~\cite{rudovic2019multi_mmql}. Given unlabeled pair data, the algorithm aggregates the results of each modality network and utilizes the aggregated results as a state for reinforcement learning. Action is defined using binary values ($0$ or $1$). If an action is $1$, input data is queried to the annotator.

\section{Methodology}
\label{section:algorithm}

\subsection{AL Scenario for Finely-Categorized ITR}
\label{subsection:scenario}
In this section, we set up an AL scenario considering the characteristics of the finely-categorized ITR dataset.
In $e$-th epoch of the scenario, AL algorithm actively selects a set of $b$ valuable images ($Q^{(e)}$) from the unpaired image dataset ($X^{(e)}$). Then an annotator provides a proper text for each image in $Q^{(e)}$, which yields a paired set ($P^{(e)}$).
Then $P^{(e)}$ is added to the set of accumulated relevant pairs $P_a^{(e)}$, whereas $Q^{(e)}$ is subtracted from $X^{(e)}$, resulting in $P_a^{(e+1)}=P_a^{(e)} \cup P^{(e)}$ and $X^{(e+1)}=X^{(e)}\setminus Q^{(e)}$, respectively.
Then a retrieval model $\mathcal{M}$ is trained by the accumulated paired dataset $P_a^{(e+1)}$. 

To represent initial states, we set the initial epoch index to zero, i.e., $e=0$. Algorithm~\ref{alg:active_scenario} describes a detailed procedure of the proposed AL scenario for ITR.
In the scenario, an unpaired image dataset is given and the annotator provides paired texts for the queried images. 
The reverse scenario of Algorithm~\ref{alg:active_scenario} can be defined in a similar manner, where an unpaired text dataset is given and the annotator provides images for the queried texts.
\begin{algorithm} [t]
\caption{AL scenario}
\label{alg:active_scenario}
\begin{flushleft}
{\bf Input:} \\
$\mathcal{M}^{(0)}$: Initial retrieval model\\
$P_a^{(0)}$: Initial accumulated paired dataset\\
$X^{(0)}$: Initial unpaired image dataset\\
$b$: The number of images to be selected at each epoch\\
$E$: Maximum epoch\\
{\bf Functions:}\\
$train(\mathcal{M}, P_a)$: Train model $\mathcal{M}$ with dataset $P_a$\\
$AL(X, b, \cdot)$: Actively selects $b$ images from $X$\\
$Annotator(Q)$: Annotate images in Q\\
{\bf Procedure:}\\
\end{flushleft}
\begin{algorithmic}[1]
\STATE $\mathcal{M}  \gets train(\mathcal{M}^{(0)}, P_a^{(0)})$
\FOR{$e=0$ to $E-1$}
    \STATE \hspace{-0.25cm}\textsc{\# Active sample selection}   
    \STATE $Q^{(e)} = \{x_i\}_{i=1}^{b} = AL(X^{(e)}, b, \cdot)$; Algorithm~\ref{alg:ALalgorithm}
    \STATE $P^{(e)} = \{(x_i, t_i)\}_{i=1}^{b} = Annotator(Q^{(e)})$
    \STATE $P_a^{(e+1)}=P_a^{(e)}\cup P^{(e)}$
    \STATE $X^{(e+1)}=X^{(e)}\setminus Q^{(e)}$
    \STATE \hspace{-0.25cm}\textsc{\# Evaluation}   
    \STATE $\mathcal{M}\gets train(\mathcal{M}^{(0)},P_a^{(e+1)})$
\ENDFOR
\STATE \textbf{return} $\mathcal{M}, P_a^{(E)}$
\end{algorithmic}
\vspace{-0.1cm}
\end{algorithm}


\paragraph{Differences from existing AL scenarios}
Existing AL scenarios (including both single-modal and multi-modal tasks) assume that \textit{unlabeled} data are given. 
Unlabeled data are feed-forwarded into the target model and then  inference results for unlabeled data are obtained from the target model.
In contrast, our AL scenario assumes that \textit{unpaired} data are given.
Unpaired data could not be feed-forwarded into the target ITR model, thus inference results for unpaired data cannot be obtained.
Table~\ref{tabular:al_scenario_difference} summarizes the differences between existing and our scenario.
For this reason, lots of existing AL algorithms relying on the inference results such as Entropy sampling~\cite{1948_shannon_communication}, CAL~\cite{margatina2021active}, DBAL~\cite{2017_active_dropout_bayesian}, BADGE~\cite{ash2019BADGE}, and LLAL~\cite{2019_learning_loss} could not be applied to our scenario. 
Therefore, in the following sections, we propose the AL algorithm suitable for our AL scenario.
\begin{table}[t]
\centering
        \begin{tabular}[h]{l|ccc}
            \toprule
            AL scenario & Given & Query & Infer\\
            \midrule
            Single-modal & $x$ or $t$ & $l$ & \ding{51} \\
            Multi-modal (exist) & $(x,t)$ & $l$ & \ding{51} \\
            \midrule
            Multi-modal (ours) & $x$ or $t$ & $t$ or $x$ & \ding{55} \\
            \bottomrule
        \end{tabular}\\
\caption{Difference among AL scenarios. $x,t,l$ denotes image data, text data, and label for given data, respectively. \textbf{Given}: Unlabeled or unpaired data that should be annotated. \textbf{Query}: Factors queried to the annotator. \textbf{Infer}: Whether inference results for the given data can be obtained.
}
\label{tabular:al_scenario_difference}
\vspace{-0.2cm}
\end{table}


\subsection{Key Concept of Proposed AL Algorithm}
\label{subsection:keyidea}
In our AL scenario, AL algorithm selects unpaired images that are expected to largely improve the performance of a model.
To this end, our key idea is to select unpaired images that yield a large training loss at the training phase.
Samples yielding a large loss can be regarded as hard samples to the current model. When using the same number of training samples, the model trained with the hard samples can achieve better performance than the model trained with randomly selected samples.
In our AL algorithm, we employ the triplet ranking loss modified to emphasize the hard negative samples~\cite{2017_itr_vsepp}. 
Based on the characteristics of the triplet ranking loss, we define conditions that an image is determined as a hard negative image for a text in the paired dataset.
Then we propose a scoring function that measures the `hard negativeness' of each unpaired image sample for the texts in the paired dataset accumulated during AL.
Finally, we propose the AL algorithm that selects the samples in the order of the highest score.

\subsection{Loss Function for Training ITR Model}
To perform ITR, a retrieval model is trained to yield high similarity for a relevant pair, and low similarity for an irrelevant pair.
To this end, the triplet ranking loss is usually adopted for the model training~\cite{chechik2010large, frome2007learning}. 
Especially, we consider the max of hinges loss function~\cite{2017_itr_vsepp} that emphasizes the hard negative samples, i.e., we penalize only the most irrelevant pair which gives the highest similarity score.
For a relevant pair $(x, t)$ for image $x$ and text $t$, the max of hinges loss is defined by
\begin{equation}
\begin{aligned}
    l(x,t) = & \max_{t'}[\alpha + s(x, t') - s(x,t)]_{+} \\
             & + \max_{x'}[\alpha + s(x', t) - s(x,t)]_{+}\\
           = & [\alpha + s(x, t^{(-)}) - s(x,t)]_{+}\\
             & + [\alpha + s(x^{(-)}, t) - s(x,t)]_{+},
 \end{aligned}
\label{eq:max_hinge_loss}
\end{equation}
where $x'$ (or $t'$) represents any one image (or text) except for $x$ (or $t$) in the training mini-batch.
The hard negative image (or text) is denoted as $x^{(-)} = \text{arg}\max_{x'}{s(x',t)}$ ($t^{(-)} = \text{arg}\max_{t'}{s(x,t')}$).
$s(x,t)$ denotes the cosine similarity between $x$ and $t$, and $[x]_{+}$ denotes the hinge function as: $[x]_{+}=\text{max}(x,0)$. $\alpha$ is a margin for the ranking loss.

\subsection{Proposed Hard Negative Conditions}\label{subsection:condition}
According to Eq.~\ref{eq:max_hinge_loss}, the hard negative image $x^{(-)}$ makes the loss large
and so $x^{(-)}$ can be chosen as a valuable image for AL. 
For a given relevant pair sample $(x,t)$, we can obtain the hard negative sample $x^{(-)}$ from Eq.~\ref{eq:max_hinge_loss}. However, in our AL scenario, because only unpaired samples are given, we cannot obtain the hard negative sample $x^{(-)}$ from Eq.~\ref{eq:max_hinge_loss}. 
To circumvent this, 
we propose an approximate condition to choose the hard negative unpaired image for a certain text in the given paired data.
For convenience, the condition is referred to as `\textit{hard negative condition}'.

To establish the hard negative condition, let $Z$ (or $T$) be a set of images (or texts) within the set of accumulated relevant pairs $P_a$.
When describing the procedure in each epoch, we omit the superscript `(e)' for simplicity. $x_i$ denotes the $i$-th image in $X$. Additionally, $z_j$ (or $t_j$) as the $j$-th sample in $Z$ (or $T$). Note that the same subscript for $z$ and $t$ means that they are relevant.
Then, we define the hard negative condition of $x_i$ regarding $t_j$ as below.

\vspace{0.1cm}
{\bf Hard negative condition:} {$x_i \in X$ is determined as the hard negative image of $t_j$ if $\;s(x_i, t_j) > \xi_j(t_j)$, where $\xi_j(t_j)$ is a threshold to be designed depending on $t_j$.}
\vspace{0.1cm}

$\xi_j$ can be designed in several ways. The first way is called `Full-batch$+$Top-$1$' condition to design a threshold $\xi_j$ to choose $x_i$ such that it should have higher similarity with $t_j$ than all other images $z_l$ in $Z\setminus\{z_j\}$. To this end, we design $\xi_j$ as
\begin{equation}
\begin{aligned}
    \xi_j = \max_{z_l} {\{s(z_l, t_j) \;|\; z_l\in Z\setminus\{z_j\}\}},
\end{aligned}
\label{eq:condition_full_batch}
\end{equation}
where
$Z$ (full-batch) is the image set of $P_a$ and so its size is large, which requires heavy computation when using Eq.~\ref{eq:condition_full_batch}. 

To reduce the computation, we design 
a relaxed version (called  `Mini-batch$+$Top-$1$' condition) of Eq.~\ref{eq:condition_full_batch} using a small subset (mini-batch) of $Z$ as 
\begin{equation}
\begin{aligned}
    \xi_j = \max_{z_l}{\{s(z_l, t_j) \;|\; z_l\in Z_s\}},
\end{aligned}
\label{eq:condition_mini_batch}
\end{equation}
where $Z_s$ is a randomly chosen subset of $Z\setminus\{z_j\}$. We also define $T_s \subset T$ as a text subset corresponded to $Z_s$.

Another way for relaxing Eq.~\ref{eq:condition_full_batch} is to replace the max function with the top-k max function which returns $k$-th largest value (called `Full(or Mini)-batch$+$Top-$k$' condition), that is, 
$\xi_j$ is designed as 
\begin{equation}
\begin{aligned}
    \xi_j = k^{\text{th}}\max_{z_l} {\{s(z_l, t_j) \;|\; z_l\in Z\setminus\{z_j\}(\text{or }Z_s) \}}.
\end{aligned}
\label{eq:condition_topk}
\end{equation}

In the ablation study, we validate the effectiveness of the threshold designs and choose one considering both computation and performance.


\begin{algorithm} [t]
\caption{Proposed AL Algorithm}
\label{alg:ALalgorithm}
\begin{flushleft}
{\bf Input:} \\
$\mathcal{M}$: Retrieval model\\
$P_a$: Accumulated paired dataset\\
$X$: Unpaired image dataset\\
$b$: The number of images to be selected at each epoch\\
{\bf Functions:}\\
$s(x, t)$: Calculate similarity between image $x$ and text $t$ \\
$[x]_{+}$: Return maximum value between $x$ and $0$\\
$\textbf{1}(c)$: Return $1$ / $0$ if condition $c$ is true / false\\
{\bf Procedure:}\\
\end{flushleft}
\begin{algorithmic}[1]
    \STATE $n_l=|P_a|$ 
    \STATE $n_u=|X|$ 
    \STATE Split $P_a=\{(z_j, t_j)\}_{j=1}^{n_l}$ into \newline $Z=\{z_j\}_{j=1}^{n_l}, T=\{t_j\}_{j=1}^{n_l}$
    \STATE \textsc{\# Calculate Threshold}
    \FOR{$j=1$ to $n_l$} 
        \STATE $t_j \gets j$-th sample of $T$
        \STATE $z_j \gets j$-th sample of $Z$
        \STATE $\xi_j = \max{\{s(z_l, t_j) \;|\; z_l\in Z\setminus\{z_j\}\}}$
    \ENDFOR
    
    \STATE \textsc{\# Calculate Score}
    \FOR{$i=1$ to $n_u$}
        \FOR{$j=1$ to $n_l$}
            \STATE $t_j \leftarrow j$-th sample of $T$
            \STATE $w_{ij} = \big[ s(x_i, t_j) - \xi_j \big]_{+}$
        \ENDFOR
        \STATE $h_i = \sum_{j=1}^{n_l}{w_{ij} \cdot \textbf{1}(s(x_i, t_j ) > \xi_j)}$
    \ENDFOR
    \STATE $Q \gets$ select $b$ images with the highest $h_i$ from $X$ 
    
    \STATE \textbf{Return} $Q$
\end{algorithmic}
\end{algorithm}

\begin{figure*}[ht]
\centering 
\includegraphics[width=1.00\linewidth]{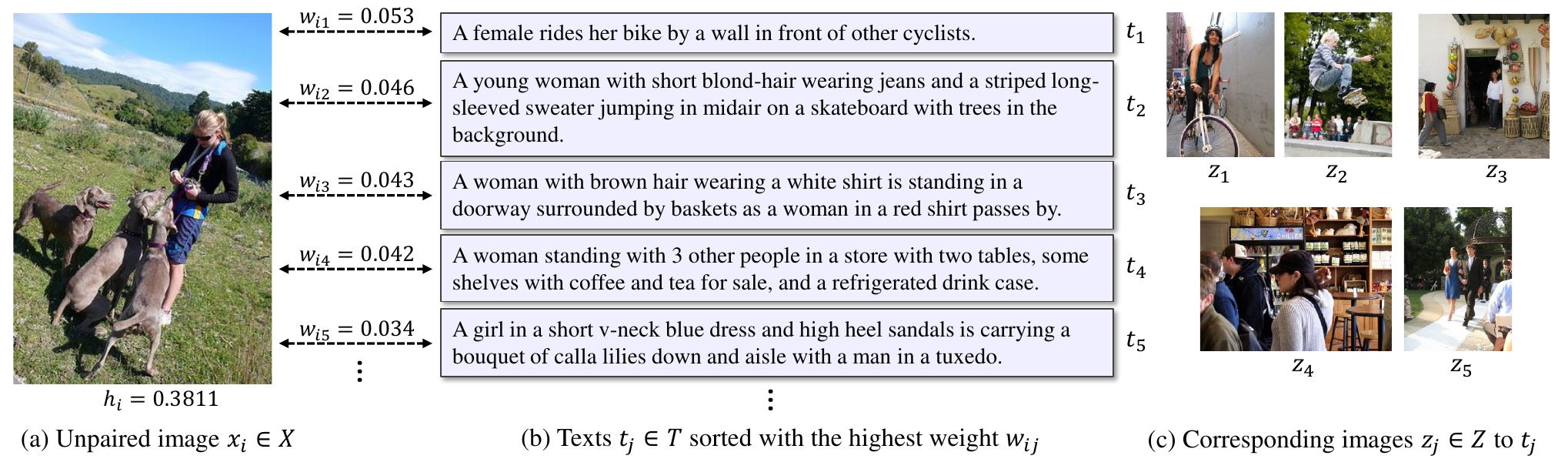}
    \caption{Example of an image selected by the proposed AL algorithm. 
    \textbf{(a)} For an unpaired image $x_i$,
    \textbf{(b)} the algorithm calculates an aggregation weight $w_{ij}$ for text data $t_j\in T$ and $x_i$, according to the designed threshold for the hard negative condition. Then a score $h_i$ for $x_i$ can be obtained by sum $w_{ij}$ along $j$.
    \textbf{(c)} Corresponding images for texts in (b). 
    }
\label{fig:experiment_qualitative}
\end{figure*}

\subsection{Proposed AL Algorithm}\label{subsection:algorithm}
Utilizing the hard negative condition, we propose a scoring function that measures the `hard negativeness' of each unpaired image sample for the texts in the paired dataset, accumulated during AL.
The proposed scoring function of $x_i$ counts the number of $t_j \in T$ for which $x_i$ satisfies the hard negative condition.
To this end, the hard negativeness score function $h_i(x_i)$ is defined by 
\begin{equation}
\begin{aligned}
    h_i(x_i) = \sum_{t_j \in T}{w_{ij} \cdot \textbf{1}(s(x_i, t_j ) > \xi_j)},
\end{aligned}
\label{eq:score}
\end{equation}
where $\textbf{1}(\cdot)$ is indicator function that returns $1$ if the input condition is satisfied, otherwise returns $0$. $w_{ij}$ is an aggregation weight for $x_i$ and $t_j$.
For the mini-batch condition, $T$ in Eq.~\ref{eq:score} is replaced with $T_s$.

Note that when $w_{ij}=1$ for all $t_j$, then $h_i$ merely counts the number of text for which $x_i$ satisfies the hard negative condition. This is referred to as \textbf{Counting} weight. On the other hand, we can suppose to give more weight to the harder negatives as follows
\begin{equation}
w_{ij} = \big[ s(x_i, t_j) - \xi_j \big]_{+}. 
\label{Surplus}
\end{equation}
In this case, $w_{ij}$ aims to give weights for $(x_i, t_j)$ such that $s(x_i, t_j) > {\xi}_j$. This is referred to as \textbf{Surplus} weight.

Finally, the proposed AL algorithm selects $b$ images from $X$, in the order of the highest score. Algorithm~\ref{alg:ALalgorithm} describes the detailed procedure of the proposed AL algorithm with the combination of Full-batch and Top-1 conditions and Surplus weight.
Figure~\ref{fig:experiment_qualitative} shows a procedure of score calculation for an unpaired image example.

\section{Experiments}
\label{section:experiment}

\subsection{Experimental Settings}
\paragraph{Dataset}
We have evaluated the proposed algorithm on MS-COCO \cite{2014_mscoco} and Flickr30K~\cite{2014_flickr} datasets, which are popular fine-grained categorized benchmarks for ITR.
Detailed configuration and preprocessing methods for datasets are described in the Appendix A of the supplementary.

\paragraph{Retrieval Model and Training Scheme}
For validation of the proposed method, we employed Iterative Matching with Recurrent Attention Memory network (IMRAM)~\cite{2020_itr_imram} a our retrieval model which is one of the state-of-the-art ITR.
For computational efficiency, we used Text-IMRAM. The hyper-parameters were set following the IMRAM paper.

At each epoch of the proposed scenario, we trained a retrieval model from scratch, with Adam optimizer~\cite{2014_adam} during $40$ epochs. Learning rate was initially set to $0.0002$ and decayed to $0.00002$ at $20$ epoch. We performed validation during last $10$ epoch, and chose the model with the best validation performance.

\paragraph{AL Settings}
Randomly selected $30\%$ of the entire paired data were assigned to $P_a^{(0)}$.
Then the remaining $70\%$ images were assigned to $X^{(0)}$.
We set a maximum epoch of the AL scenario to $E=3$ and $b$ to $5\%$ of the cardinality of the entire dataset.
Therefore, after completing the scenario, $|P_a^{(E)}|$ becomes $45\%$ of the cardinality of the entire dataset.

\paragraph{Hyper-parameters} 
Based on the ablation study, we determined the hyper-parameters as follows.
For Flickr30K, the threshold $\xi$ was calculated with  Full-batch $+$ Top-$1$ condition in Eq. \ref{eq:condition_full_batch}. For MS-COCO, we calculated $\xi$ with Mini-batch $+$ Top-$1$ condition in Eq. \ref{eq:condition_mini_batch} to reduce the computational complexity. For both datasets, the aggregation weight $w$ was calculated by the Surplus weight in Eq. \ref{Surplus}.

\paragraph{Evaluation}
We evaluated a retrieval model on two tasks. In an Image Retrieval task, a model retrieves images given a text query. In a Text Retrieval task, a model retrieves texts given an image query.
Performance was measured by Recall at $K$ (R@$K$) metric. 
We evaluated the model with $K=1$.

For each epoch of the scenario, we trained the model from scratch with the accumulated paired dataset ($P_a^{(e+1)}$ for $e$-th epoch) and reported the test performance of the trained model.
For MS-COCO, the model was validated by both testing on full 5K test images (COCO-5K) and averaging the results over five subsets of 1K test images (COCO-1K).

\begin{table}[t]
\centering
        \begin{tabular}[h]{c|c|c}
            \toprule
            Condition & Weight & R@$1$-sum \\
            \midrule
            \multirow{2}{*}{Full-batch + Top-1} 
            & Surplus weight& \textbf{298.5}\\
            & Count weight & 296.6\\
            \multirow{2}{*}{Full-batch + Top-5} 
            & Surplus weight & 293.5\\
            & Count weight   & 294.8\\
            \multirow{2}{*}{Full-batch + Top-10} 
            & Surplus weight & 294.2\\
            & Count weight   & 296.8\\
     
            \midrule
            
            \multirow{2}{*}{Mini-batch + Top-1} 
            & Surplus weight & \textbf{298.6}\\
            & Count weight   & 296.3\\
            \multirow{2}{*}{Mini-batch + Top-5} 
            & Surplus weight & 296.1\\
            & Count weight   & 296.4\\
            \multirow{2}{*}{Mini-batch + Top-10} 
            & Surplus weight & 292.9\\
            & Count weight   & 293.8\\
            \bottomrule
        \end{tabular}
\caption{
R@$1$-sum performance of the proposed AL algorithm, according to the hard negative condition and the aggregation weight for the Flickr30K.}
\label{tabular:ablation_condition_weight}
\end{table}

\subsection{Ablation and Self Study}
\label{section:ablation}

\begin{table}
\centering
        \begin{tabular}[h]{c|ccc}
            \toprule
            \multirow{2}{*}{Condition} & \multicolumn{3}{c}{\% of Hard neg. Images}\\
            & $e=0$ & $e=1$ & $e=2$\\
            \midrule
            Full-batch + Top-$1$  & 55.04 & 44.85 & 39.72\\
            Full-batch + Top-$5$  & 92.64 & 89.85 & 88.17\\
            Full-batch + Top-$10$ & 98.27 & 97.51 & 96.98\\
            \midrule
            Mini-batch + Top-$1$  & 56.23 & 49.71 & 47.11\\
            Mini-batch + Top-$5$  & 95.26 & 93.94 & 92.76\\
            Mini-batch + Top-$10$ & 99.24 & 99.14 & 99.19\\
            \bottomrule
        \end{tabular}
\caption{Percentage of the hard negative unpaired images in the unpaired image dataset depending on the hard negative condition.
}
\label{tabular:ablation_relaxation}
\end{table}

\subsubsection{Validation on Hyper-parameters}
\label{subsection:hyperparam_ablation}
We evaluated the proposed algorithm with various hard negative conditions and aggregation weights for Flickr30K.
To evaluate the overall performance over epoch, in Table~\ref{tabular:ablation_condition_weight}, we reported R@$1$-sum metric that sums all up R@$1$ performances over epochs for both text retrieval and image retrieval tasks for Flickr30K.
In the Appendix B of the supplementary, R@$1$ performance at each epoch of AL scenario are provided.

According to the results in Table~\ref{tabular:ablation_condition_weight}, 
the combination of Top-$1$ condition and Surplus weight achieves the best R@$1$-sum performance by $298.5$ and $298.6$, about $2\sim6$ better than the other combinations.
Therefore, we mainly considered Top-$1$ condition and Surplus weight as our default setting.
Unless otherwise specified for the top-k condition and the aggregation weight, the Full-batch / Mini-batch version of the proposed algorithm indicates the algorithm to which a combination of Full-batch / Mini-batch and Top-$1$ condition and Surplus weight are applied.

Each of Full-batch and Mini-batch versions has its own advantages and disadvantages.
Full-batch version has advantages in performance.
For R@$1$-sum, R@$5$-sum, and R@$10$-sum values 
(R@$5$ and R@$10$ results are provided in the Appendix B.), Full batch version achieves  $298.5 / 520.0 / 602.9$, whereas Mini-batch version achieves $298.6 / 520.2 / 600.7$, respectively.
Though both show similar R@$1$-sum and R@$5$-sum performance, the Full-batch version is slightly better in the R@$10$-sum. On the other hand, Mini-batch version is computationally efficient. 
Mini-batch version requires similarity calculations of $|Z_s|\cdot|T_s| + |X|\cdot|T_s|$, whereas Full-batch version requires $|Z|\cdot|T| + |X|\cdot|T|$.
In our experimental setting on Flickr30K, $|Z_s|=|T_s|=2560$ is considerably smaller than $|Z|=|T|\in[8700, 13050]$. (For MS-COCO, $|Z|\in[24834, 37252]$)
Therefore, a trade-off in performance and computational cost can be negotiated between Full and Mini-batch versions.

\subsubsection{Validation on Relaxed Condition}
\label{subsection:validimage_ablation}
In the proposed algorithm,
we also suggested relaxed conditions
including Mini-batch condition (Eq.~\ref{eq:condition_mini_batch}) and Top-k condition (Eq.~\ref{eq:condition_topk}).
To validate the relaxation effect of proposed conditions, we chose hard negative unpaired images satisfying the relaxed condition for at least one text in the paired dataset.
Then we compared the ratio of the hard negative images in the unpaired image dataset with those of the non-relaxed conditions.
High ratio means that the relaxation effect is large. 

Table~\ref{tabular:ablation_relaxation} shows the ratio of hard negative images at each epoch of the AL scenario, depending on hard negative conditions for Flickr30K. 
According to Table~\ref{tabular:ablation_relaxation}, the ratio of hard negative images increases when Mini-batch condition is applied and $k$ of Top-k condition increases.
This implies that the combination of Mini-batch and Top-k conditions gives the largest relaxation effect.

It is interesting to note that the ratio of the hard negative images decreases as the AL scenario progresses. 
The more training paired data accumulated as the AL scenario progresses, any unpaired image is more likely to be similar to the images in the accumulated paired dataset.
In addition, the retrieval model is trained to predict low similarity between texts and negative images in the accumulated paired dataset.
Therefore, the retrieval model is likely to predict low similarity between any unpaired image and texts in the accumulated dataset.
In consequence, the number of unpaired images that satisfy the condition will decrease.

\begin{figure}[t!]
\centering 
\includegraphics[width=1.0\linewidth]{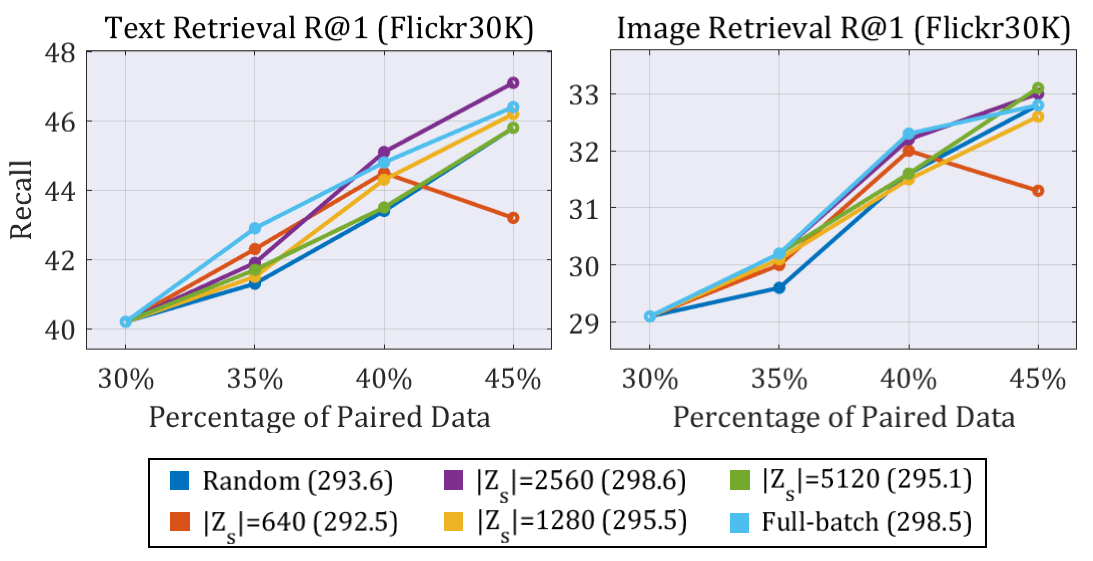}
    \caption{
    Performance of the proposed algorithm with Mini-batch version, according to the cardinality of $Z_s$. 
    }
\label{fig:ablation_batchsize}
\vspace{-0.5cm}
\end{figure}
\begin{figure}[t]
\centering 
\includegraphics[width=1.00\linewidth]{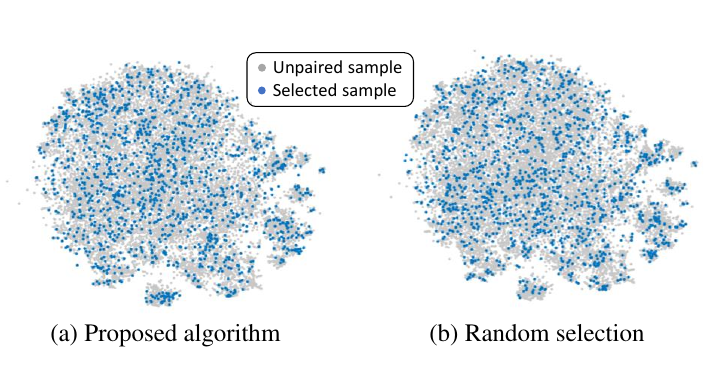}
    \caption{Visualization of images selected by (a) the proposed algorithm and (b) the random selection in the tSNE embedding space.}
\label{fig:appendix_distribution}
\vspace{-0.5cm}
\end{figure}

\subsubsection{$|Z_s|$ for Mini-batch Condition}
\label{subsection:batchsize_ablation}
For the mini-batch version of the proposed algorithm, the cardinality of subset $Z_s$, $|Z_s|$, needs to be determined. To validate the effect of $|Z_s|$, we evaluated the Mini-batch version algorithm by increasing $|Z_s|$ from $640$ to $5120$.

Figure~\ref{fig:ablation_batchsize} shows R@$1$ performance of Mini-batch version with various $|Z_s|$. The number in the legend indicates the R@$1$-sum performance of each algorithm.
When $|Z_s| \geq 1280$, the mini-batch version shows better R@$1$-sum performance than random selection. The best R@$1$-sum performance is achieved when $|Z_s|=2560$, which is equivalent to $40$ times of the training mini-batch size and 20\% of the full-batch version.
We fixed $|Z_s|=2560$ for Mini-batch version during the other experiments.

\begin{figure*}[ht!]
\centering 
\includegraphics[width=0.98\linewidth]{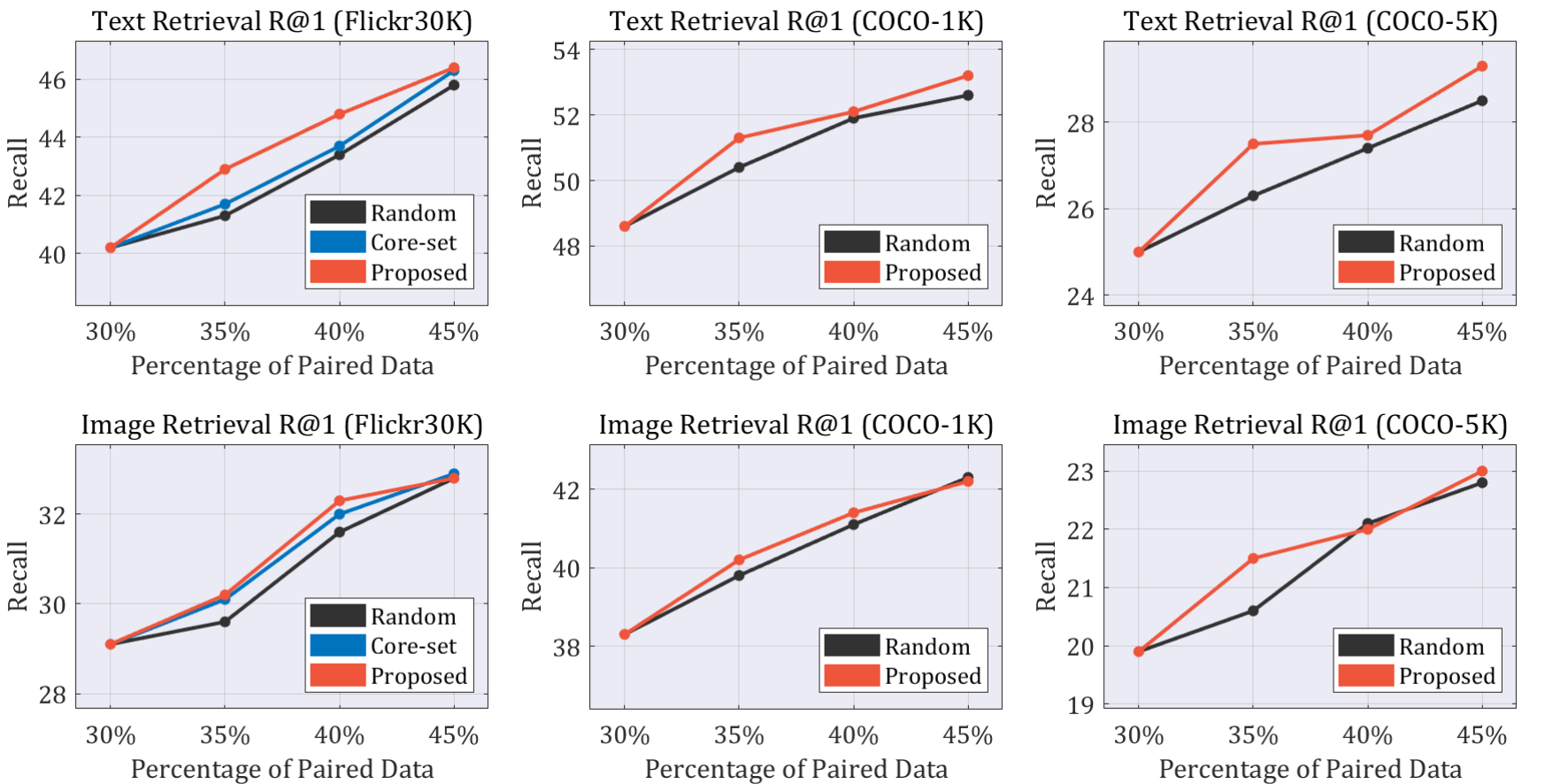}
    \vspace{0.2cm}
    \caption{Evaluation results on Flickr30K and MS-COCO. Each graph shows R@$1$ performance at each epoch of the AL scenario. $x$-axis represents the ratio of paired data to the entire dataset and $y$-axis denotes the R@$1$ performance.}
\label{fig:experiment_r1}
\end{figure*}

\subsubsection{Distribution of Selected Samples}
Several deep AL studies~\cite{2018_active_coreset,kirsch2019batchbald} have pointed out the problem that the score-based AL algorithms select biased and overlapped samples.
To observe the tendency of samples selected by the proposed algorithm, we visualized the selected samples in tSNE embedding space.
Figure~\ref{fig:appendix_distribution} shows the visualization results of the samples selected by the proposed method (left) and random selection (right). Blue and gray circles represent the selected and non-selected unpaired images, respectively.
The selected images extracted by both our and random methods are evenly distributed over the entire region, which shows that our results also are not biased to a region, along with achieving meaningful improvement of performance.

\subsection{Comparisons}
\label{subsection:experiment_results}
Since our AL scenario is the new one, as we mentioned in Methodology section, most of the existing AL algorithms are difficult to be applied to our scenario. Hence we compared the proposed method with the random selection and Core-set algorithm modified for our AL scenario~\cite{2018_active_coreset}.
Detailed implementation of modified Core-set is provided in the Appendix C of the supplementary. For MS-COCO, Core-set was not compared due to the GPU memory limitation.

Figure~\ref{fig:experiment_r1} shows the evaluation results.
We can see that the proposed method achieves the best performance in most epochs, compared to both random baseline and Core-set.
When performing image-text retrieval, only the similarity between image and text is considered.
Thus, the proposed method that selects images by considering the similarity between image-text seems to select more valuable samples rather than the Core-set method that considers only the relationship between images.
In fact, as shown in Figure~\ref{fig:experiment_qualitative}, the images corresponding to texts having the highest similarity to $x_i$ are not very similar to $x_i$.

An interesting point is that, at the first epoch (when the percentage of paired data is 35\%), the proposed algorithm achieves much better R@$1$ performance than the other algorithms by $[0.4\sim1.6]$. 
Since the proposed algorithm selects hard negative images for the retrieval model, the hard negative images decrease as epochs progress. Thus the proposed algorithm needs to select harder negative images than the previously selected hard negative images.
However, it is much more difficult to select hard negative images in later epochs of AL scenario.
Therefore, the proposed method shows especially high performance in the first epoch.

\section{Conclusion}
\label{section:Conclusion}
In this paper, 
first, we have suggested an AL scenario for finely-categorized ITR where unpaired images are given and the annotator provides their corresponding paired texts.
Then we have developed our own AL algorithm that chooses unpaired samples that are expected to yield a large training loss, especially max of hinges loss~\cite{2017_itr_vsepp}.
The key components of our AL algorithm are (1) hard negative conditions to mine the hard negative images for constructing new paired data; (2) the scoring formula that weighs the number of texts satisfying the hard negative conditions, which is used as the criterion to determine the hard negative images.
We demonstrated the effectiveness of the proposed algorithm 
through extensive experiments for ablation and self studies, and comparisons with the existing works on Flickr30K and MS-COCO.

\clearpage

\setcounter{figure}{0}
\renewcommand{\thefigure}{A\arabic{figure}}

\setcounter{table}{0}
\renewcommand\thetable{A\arabic{table}}
\setcounter{secnumdepth}{0}

\section{Appendix}

\subsection{A. Dataset Configuration and Preprocessing}
We have evaluated the proposed algorithm on MS-COCO \cite{2014_mscoco} and Flickr30K~\cite{2014_flickr} datasets.

\paragraph{Configuration}
\textbf{MS-COCO} contains $82,783$ training images and $40,504$ validation images. Each image has five captions. Following~\cite{2015_karpathy_datasplit}, we utilized only $5,000$ for validation and $5,000$ images for testing from the original validation set.
Note that several studies~\cite{2018_itr_scan, 2019_itr_camp, 2019_itr_visualsemantic, 2020_itr_imram} included the remaining $30,504$ validation images into the training set, but we did not include that because the goal of this paper is not to achieve state-of-the-art ITR performance.
\textbf{Flickr30K} contains $31,014$ images and five captions are provided for each image. Following~\cite{2015_karpathy_datasplit}, we split the dataset into $29,000$ training images, $1,014$ validation images, and $1,000$ testing images.
Since each image has five captions, we generated five positively relevant pairs.
But many recent studies~\cite{2020_itr_imram, 2018_itr_scan} assumed that each pair among the five pairs has a different category from the others although five pairs share the same image.
This data processing might not be a fatal problem for training the ITR model.
However, for AL task, the image sharing might be problematic because the same image can be selected by an AL algorithm.
Thus, we used only one caption for each image for training. On the other hand, for test and validation, we used all five captions for each image.

\paragraph{Feature Extraction}
For each image, following \cite{2018_bottomup}, we extracted $36$ local features using Faster R-CNN~\cite{2015_faster_rcnn} with ResNet-101 backbone~\cite{2016_resnet} pretrained on Visual Genome dataset~\cite{2017_visual_genome}.
Each local feature vector for image has $2048$-dimension.
For text data, following~\cite{2018_itr_scan}, each word in sentence was embedded into  $300$-dimensional vector first. Then we utilized bi-directional GRU~\cite{bahdanau2014neural} to extract final feature for each word.
Final feature vector for word has $1024$-dimension.

\subsection{B. Validation on Hyper-parameters}
Table~\ref{tabular:ablation_condition_weight_r1}, \ref{tabular:ablation_condition_weight_r5} and \ref{tabular:ablation_condition_weight_r10} provide R@$1$, R@$5$ and R@$10$ performances respectively, at each epoch of AL scenario for Flickr30K, given various combinations of the hard negative conditions and aggregation weights.

According to the results in tables, the dominant combination that achieves the best performance at every epoch does not exist. 
But the combination of Top-$1$ condition and Surplus weight achieves the best R@$1$-sum, R@$5$-sum, and R@$10$-sum  performance, about $2\sim6$ better than the other combinations.
Therefore, we mainly considered Top-$1$ condition and Surplus weight as our default setting in the main document.

\begin{table}
\centering
        \begin{tabular}[h!]{l|cc|cc|c}
            \toprule
            \multirow{2}{*}{Algorithm} & \multicolumn{2}{c|}{Text Ret.} & \multicolumn{2}{c|}{Image Ret.} & R@$1$\\
            & $e$ = $1$ & $e$ = $3$ & $e$ = $1$ & $e$ = $3$ & -sum\\
            \midrule
            Core-set-mean & 41.2 & 45.2 & 30.2 & 33.2 & 295.1\\
            Core-set-BoW  & 41.7 & 46.3 & 30.1 & 32.9 & \textbf{295.9}\\
            \bottomrule
        \end{tabular}
\caption{
Performance of Core-set depending on the feature extraction method.}
\label{tabular:ablation_coreset}
\small
\end{table}
\begin{table*}[t]
\centering
        \begin{tabular}[h]{c|c|cccc|cccc|c}
            \toprule
            \multirow{2}{*}{Condition} & \multirow{2}{*}{Weight} & \multicolumn{4}{c|}{Text Retrieval (R@$1$)} & \multicolumn{4}{c|}{Image Retrieval (R@$1$)} & \\
            && $e=0$ & $e=1$ & $e=2$ & $e=3$ & $e=0$ & $e=1$ & $e=2$ & $e=3$ & R@1-sum \\
            \midrule
            \multirow{2}{*}{Full-batch + Top-1} 
            & Surplus & 40.2 & 42.9 & 44.8 & 46.4 & 29.1 & 30.2 & 32.3 & 32.8 & \textbf{298.5}\\
            & Count & 40.2 & 42.1 & 44.7 & 45.3 & 29.1 & 29.9 & 32.0 & 33.5 & 296.6\\
            \multirow{2}{*}{Full-batch + Top-5} 
            & Surplus & 40.2 & 42.4 & 42.8 & 44.9 & 29.1 & 30.3 & 31.1 & 32.7 & 293.5\\
            & Count   & 40.2 & 41.4 & 44.0 & 45.4 & 29.1 & 30.4 & 31.6 & 32.8 & 294.8\\
            \multirow{2}{*}{Full-batch + Top-10} 
            & Surplus & 40.2 & 42.2 & 43.0 & 45.8 & 29.1 & 29.8 & 31.3 & 32.9 & 294.2\\
            & Count   & 40.2 & 41.7 & 45.0 & 46.7 & 29.1 & 29.9 & 31.4 & 32.9 & 296.8\\
     
            \midrule
            
            \multirow{2}{*}{Mini-batch + Top-1} 
            & Surplus & 40.2 & 41.9 & 45.1 & 47.1 & 29.1 & 30.2 & 32.2 & 33.0 & \textbf{298.6}\\
            & Count   & 40.2 & 42.6 & 44.7 & 45.0 & 29.1 & 30.1 & 32.1 & 32.5 & 296.3\\
            \multirow{2}{*}{Mini-batch + Top-5} 
            & Surplus & 40.2 & 41.1 & 45.0 & 45.7 & 29.1 & 30.0 & 31.9 & 33.3 & 296.1\\
            & Count   & 40.2 & 43.2 & 44.0 & 45.7 & 29.1 & 30.5 & 31.2 & 32.6 & 296.4\\
            \multirow{2}{*}{Mini-batch + Top-10} 
            & Surplus & 40.2 & 40.8 & 44.1 & 45.4 & 29.1 & 29.6 & 31.5 & 32.3 & 292.9\\
            & Count   & 40.2 & 41.2 & 44.8 & 44.6 & 29.1 & 29.9 & 31.6 & 32.6 & 293.8\\
            \bottomrule
        \end{tabular}
\caption{
R@$1$ performance of the proposed AL algorithm at each epoch of AL scenario, according to the hard negative condition and the aggregation weight for the Flickr30K.}
\label{tabular:ablation_condition_weight_r1}
\end{table*}

\begin{table*}[!ht]
\centering
        \begin{tabular}[h]{c|c|cccc|cccc|c}
            \toprule
            \multirow{2}{*}{Condition} & \multirow{2}{*}{Weight} & \multicolumn{4}{c|}{Text Retrieval (R@$5$)} & \multicolumn{4}{c|}{Image Retrieval (R@$5$)} & \\
            && $e=0$ & $e=1$ & $e=2$ & $e=3$ & $e=0$ & $e=1$ & $e=2$ & $e=3$ & R@5-sum \\
            \midrule
            \multirow{2}{*}{Full-batch + Top-1} 
            & Surplus & 70.2 & 71.0 & 73.9 & 75.1 & 54.3 & 57.0 & 58.7 & 59.9 & \textbf{520.0} \\
            & Count   & 70.2 & 70.9 & 73.3 & 75.0 & 54.3 & 56.2 & 58.2 & 59.8 & 517.8 \\
            \multirow{2}{*}{Full-batch + Top-5} 
            & Surplus & 70.2 & 70.5 & 74.1 & 73.5 & 54.3 & 56.1 & 57.7 & 59.5 & 515.7 \\
            & Count   & 70.2 & 70.7 & 73.3 & 74.9 & 54.3 & 56.3 & 58.5 & 59.1 & 517.2 \\
            \multirow{2}{*}{Full-batch + Top-10} 
            & Surplus & 70.2 & 70.9 & 72.6 & 74.7 & 54.3 & 55.7 & 57.6 & 59.2 & 515.1 \\
            & Count   & 70.2 & 54.3 & 74.3 & 74.4 & 54.3 & 56.1 & 58.0 & 59.2 & 516.7 \\
     
            \midrule
            
            \multirow{2}{*}{Mini-batch + Top-1} 
            & Surplus & 70.2 & 72.2 & 73.6 & 75.6 & 54.3 & 56.1 & 58.6 & 59.7 & \textbf{520.2} \\
            & Count   & 70.2 & 71.3 & 73.3 & 74.7 & 54.3 & 56.6 & 58.6 & 59.8 & 518.6 \\
            \multirow{2}{*}{Mini-batch + Top-5} 
            & Surplus & 70.2 & 70.8 & 73.2 & 74.2 & 54.3 & 56.3 & 57.6 & 59.1 & 515.6 \\
            & Count   & 70.2 & 70.8 & 73.2 & 74.2 & 54.3 & 56.3 & 57.6 & 59.1 & 515.9 \\
            \multirow{2}{*}{Mini-batch + Top-10} 
            & Surplus & 70.2 & 71.9 & 72.9 & 74.4 & 54.3 & 56.0 & 57.7 & 58.9 & 516.2 \\
            & Count   & 70.2 & 70.5 & 74.1 & 74.2 & 54.3 & 56.1 & 57.8 & 58.8 & 515.9 \\
            \bottomrule
        \end{tabular}
\caption{
R@$5$ performance of the proposed AL algorithm at each epoch of AL scenario, according to the hard negative condition and the aggregation weight for the Flickr30K.}
\label{tabular:ablation_condition_weight_r5}
\end{table*}
\begin{table*}[h!]
\centering
        \begin{tabular}[ht]{c|c|cccc|cccc|c}
            \toprule
            \multirow{2}{*}{Condition} & \multirow{2}{*}{Weight} & \multicolumn{4}{c|}{Text Retrieval (R@$10$)} & \multicolumn{4}{c|}{Image Retrieval (R@$10$)} & \\
            && $e=0$ & $e=1$ & $e=2$ & $e=3$ & $e=0$ & $e=1$ & $e=2$ & $e=3$ & R@10-sum \\
            \midrule
            \multirow{2}{*}{Full-batch + Top-1} 
            & Surplus & 80.2 & 81.6 & 83.4 & 84.4 & 65.5 & 67.8 & 69.2 & 70.9 & \textbf{602.9} \\
            & Count   & 80.2 & 81.0 & 82.9 & 83.8 & 65.5 & 67.3 & 68.7 & 70.4 & 599.6 \\
            \multirow{2}{*}{Full-batch + Top-5} 
            & Surplus & 80.2 & 80.6 & 82.4 & 83.8 & 65.5 & 67.2 & 68.7 & 70.2 & 598.5 \\
            & Count   & 80.2 & 80.9 & 82.9 & 83.7 & 65.5 & 67.0 & 68.6 & 69.9 & 598.5 \\
            \multirow{2}{*}{Full-batch + Top-10} 
            & Surplus & 80.2 & 80.6 & 81.3 & 84.1 & 65.5 & 66.9 & 68.5 & 70.2 & 597.2 \\
            & Count   & 80.2 & 81.1 & 83.4 & 83.4 & 65.5 & 67.2 & 69.0 & 69.8 & 599.4 \\
     
            \midrule

            \multirow{2}{*}{Mini-batch + Top-1} 
            & Surplus & 80.2 & 81.3 & 83.3 & 84.5 & 65.5 & 66.9 & 69.2 & 69.9 & \textbf{600.7} \\
            & Count   & 80.2 & 81.9 & 82.5 & 84.5 & 65.5 & 67.1 & 68.7 & 70.0 & 600.3 \\
            \multirow{2}{*}{Mini-batch + Top-5} 
            & Surplus & 80.2 & 81.2 & 83.2 & 84.2 & 65.5 & 66.8 & 68.4 & 69.6 & 598.8 \\
            & Count   & 80.2 & 80.8 & 82.9 & 84.5 & 65.5 & 67.3 & 68.5 & 70.0 & 599.5 \\
            \multirow{2}{*}{Mini-batch + Top-10} 
            & Surplus & 80.2 & 81.8 & 81.5 & 83.3 & 65.5 & 66.6 & 68.4 & 69.4 & 596.5 \\
            & Count   & 80.2 & 80.8 & 84.1 & 83.8 & 65.5 & 67.1 & 68.3 & 69.6 & 599.2 \\
            \bottomrule
        \end{tabular}
\caption{
R@$10$ performance of the proposed AL algorithm at each epoch of AL scenario, according to the hard negative condition and the aggregation weight for the Flickr30K.}
\label{tabular:ablation_condition_weight_r10}
\end{table*}


\subsection{C. Implementation Details for Core-set}
Core-set algorithm~\cite{2018_active_coreset} extracts one global feature vector from an image sample, and then utilizes distances between any two feature vectors. However, IMRAM model extracts local feature vectors from 36 local regions in an image. Therefore, in order to apply Core-set method to our setting, it is necessary to extract one global feature vector for an image.

We obtain the global feature vector via two methods. One is to extract a 2048 dimensional global feature vector by average 36 local feature vectors extracted from an image, referred to as `Core-set-mean' in Table ~\ref{tabular:ablation_coreset}. The other one is to extract a $300$ dimensional global feature vector of which element is the number of local feature vectors belonging to a local cluster region formed by $K$-means clustering, referred to as `Core-set-BoW' in Table ~\ref{tabular:ablation_coreset}.

Table~\ref{tabular:ablation_coreset} shows the evaluation results at the first / last epoch of AL scenario and R@$1$-sum performance, when above two methods are applied to Core-set. Core-set-mean and Core-set-BoW show comparable performances, but Core-set-BoW shows slightly better performances than Core-set-mean. Therefore, we compared Core-set-BoW with ours in the experiments.

\newpage
\bibliography{egbib.bib}

\end{document}